\documentclass{article}

\usepackage{iclr2027_conference,times}
\iclrfinalcopy

\usepackage[utf8]{inputenc}
\usepackage[T1]{fontenc}
\usepackage{amsmath,amssymb}
\usepackage{booktabs}
\usepackage{graphicx}
\usepackage{subcaption}
\usepackage{microtype}
\usepackage{xcolor}
\usepackage{url}
\usepackage{hyperref}

\newcommand{\acr}{\textsc{ACR}}

\newcommand{\model}{Qwen3.6-35B-A3B}

\title{\centering Agent-Conditioned Routing:\\
Diverse Reasoning Paths Within One Frozen Model}

\author{\parbox[t]{\textwidth}{\centering
\textbf{Ananth Kalyanasundaram}\textsuperscript{\normalfont 1}\quad
\textbf{Ankush Kadu}\textsuperscript{\normalfont 1}\quad
\textbf{Athira Gopal}\textsuperscript{\normalfont 1}\quad
\textbf{Aswanth Krishnan}\textsuperscript{\normalfont 1} \\[2pt]
\normalfont\textsuperscript{1}QpiAI, India \\
\normalfont\texttt{\{ananth.k, ankush.k, athira.g, aswanth.k\}@qpiai.tech}
}}

\begin{document}

\maketitle
{\renewcommand{\thefootnote}{}%
\footnotetext{Correspondence to: Ananth Kalyanasundaram \texttt{<ananth.k@qpiai.tech>}.}%
\footnotetext{A preliminary version of this work accepted  at the NeurIPS 2026 Workshop on Resource-Aware Agentic AI (non-archival).}}

\begin{abstract}
Generating several solutions can expose alternative methods, but additional
trajectories often paraphrase one dominant approach.
We investigate whether the conditional capacity of a frozen sparse
mixture-of-experts (MoE) model can provide such diversity from one problem
prompt, without user-assigned roles, prescribed methods, fine-tuning, or
model copies. Agent-Conditioned Routing (\acr{})
assigns $m$ parallel trajectories persistent, untrained shared/private
expert-eligibility masks while preserving the learned, input-dependent token
router. On a screened 126-question multi-mode mathematics and physics benchmark,
with model judges reading the reported derivations in full, we count the number
of problems yielding at least two distinct primary methods among soft-correct
solutions in four trajectories. \acr{} reaches approximately twice as many
such problems as the strongest temperature setting under GLM 5.3
(37.3 versus 18.7), and 1.67 times as many under GPT-5.6 Sol
(45.00 versus 27.00), at matched output budgets. A blinded four-expert audit on a stratified sample of 30 problems also
finds more multi-method solutions with ACR on both backbones, with lower
correctness.
Matched-eligibility ablations show modest multi-method count advantages
for \acr{}, while token-resampled masks attain higher correctness. An informed
sequential prompt elicits more diversity still, using previous solutions
and explicit requests for alternative methods. \acr{} adds about 160~KiB of mask state. In a three-problem
microbenchmark capped at 1,024 tokens, its latency and peak VRAM usage
remain within 3\% of batched temperature sampling.
\end{abstract}

\section{Introduction}

Inference-time scaling treats additional generations as additional chances
of success, but samples that repeat one argument inherit the same blind spots.
Answer-level techniques such as self-consistency still benefit from repeated
samples on verifiable tasks~\citep{wang2023selfconsistency}, yet agreement does
not reveal whether derivations differ meaningfully. For mathematical and
scientific reasoning, we therefore study \emph{correct method diversity}:
on how many problems do four trajectories yield at least two distinct primary
solution methods with correct conclusions under a fixed generation budget?

One can seek diversity through decoding, method prompts, trained adapters,
or coordinated models. We target \emph{autonomous method diversification}:
the user supplies a problem without having to identify alternative solution
techniques in advance. We constrain expert eligibility in a frozen sparse
MoE while retaining its learned token router. MoEs expose alternative
feed-forward paths through conditional computation
\citep{shazeer2017sparsely,lepikhin2021gshard,fedus2022switch,du2022glam}.
The question is whether these constraints change the executed derivation
while preserving correctness, not merely whether they change the text.

Agent-Conditioned Routing (\acr) executes $m{=}4$ \emph{route-conditioned
trajectories} on one frozen \model{} instance in one batched request. Here
\emph{agent} denotes a route-conditioned trajectory.
All trajectories receive the same prompt and use deterministic decoding, with no
role prompts, role generation, updates, communication, or model copies.
A seeded construction
assigns each a persistent shared/private eligibility mask; the frozen learned
router still selects its top-$k$ experts within that eligible set.

Our contributions are twofold:

\begin{enumerate}
  \item \textbf{Agent-Conditioned Routing.}  We introduce a training-free
  inference mechanism that uses expert eligibility as a control for reasoning
  diversity. Shared/private masks assign fixed routing constraints during
  each trajectory's prefix, enabling diversification from one problem prompt
  without user-assigned roles, prescribed methods, or parameter updates.
  \item \textbf{A multi-mode reasoning benchmark for diversity.}  We construct
  a 126-question benchmark spanning 12 mathematics and physics categories and
  supporting multiple solution modes. The evaluation measures how often four
  trajectories yield at least two distinct primary methods among soft-correct
  solutions. Full-benchmark evaluations by two model judges are complemented
  by a blinded four-expert audit on a stratified 30-problem subset across
  three seeds. Correctness and mean UCM@4 accompany threshold counts.
\end{enumerate}

\section{Related work}
\label{sec:related}

ACR connects inference-time reasoning diversity with sparse-MoE conditional
computation, seeking distinct primary solution methods across independent
trajectories within one frozen model without specifying methods in the prompt.

\subsection{Inference-time diversity from one model}
Chain-of-thought prompting elicits intermediate reasoning, and
self-consistency samples reasoning chains before aggregating their answers
\citep{wei2022chain,wang2023selfconsistency}. Tree search, self-refinement,
and test-time compute scaling use additional inference to improve reasoning
\citep{yao2023tree,madaan2023selfrefine,snell2025testtime}. These approaches
motivate generating and evaluating alternatives, but answer-level success
does not establish that the alternatives execute different methods.

Other approaches target variation more directly. Diverse beam search
separates decoded outputs~\citep{vijayakumar2018diverse}, while diverse
prompting and Flow of Reasoning target reasoning-path diversity
\citep{lee2024diverseprompting,yu2025flow}. STaR develops reasoning ability
through training on generated rationales~\citep{zelikman2022star}.
ACR introduces a different control: it holds the prompt, weights, and
decoding rule fixed while varying expert eligibility across trajectories.
We evaluate primary executed methods among correct trajectories rather
than lexical variation or answer votes.

\subsection{Parallel-agent reasoning}
Parallel-agent systems obtain complementary outputs through multiple
reasoning processes and mechanisms for combining them. Mixture-of-Agents
and Sparse Mixture-of-Agents combine outputs across model agents
\citep{wang2025moa,li2024smoa}; multiagent debate exchanges critiques
between models~\citep{du2024debate}. Group Think and Hogwild! Inference
share partial state among concurrent workers
\citep{groupthink2025,hogwild2025}.

ACR instead places trajectory identity in routing constraints: trajectories
share one frozen backbone but generate independently, so differentiation does
not require separate weights, role prompts, or inter-agent communication.

\subsection{Expert routing as an inference-time control}
MoE research offers several ways to organize and select conditional
computation, including BASE layers, expert-choice routing, ST-MoE, Mixtral,
and DeepSeekMoE
\citep{lewis2021base,zhou2022expertchoice,zoph2022stmoe,jiang2024mixtral,dai2024deepseekmoe}.
Related interventions contrast selected and unselected experts in SCMoE
\citep{shi2024scmoe}, learn router parameters in HyperRouter
\citep{do2023hyperrouter}, or learn adapter experts and a router in Self-MoE
\citep{kang2024selfmoe}. Semantic routing probes further motivate expert
selection as an intervention point~\citep{olson2025semantic}. Recent work
finds a domain-invariant ``Standing Committee'' of heavily reused experts,
questioning clean domain specialization and motivating shared capacity
\citep{wang2026illusion}; our random shared set does not identify that committee.

The closest inference-time comparisons are RoE and Expert-Sample.
RoE resamples noisy expert routes at each token and aggregates the resulting
proposals into a prediction~\citep{zibakhsh2025roe}. Expert-Sample retains
the router's certain selections and Gumbel-resamples its uncertain tail,
seeking generation diversity while preserving answer-level performance
\citep{chen2026expertsample}. Both provide training-free ways to vary the
computation of a frozen MoE, and we evaluate them as routing-noise controls.

ACR instead assigns persistent shared/private eligibility masks, within
which the pretrained router selects experts deterministically. We evaluate
Expert-Sample in its published routing configuration and RoE as a
proposal-averaging control. Section~\ref{sec:ablations} further separates
temporal persistence from mask construction through matched-cardinality
eligibility controls.

\section{Method}
\label{sec:method}

\subsection{Persistent shared/private eligibility}

Let the frozen MoE have a routed-expert universe $\mathcal{E}$,
$|\mathcal{E}|=E$, and let its learned router emit logits
$r_\ell(h_t)\in\mathbb{R}^{E}$ at layer $\ell$ and token $t$.  We execute $m$
route-conditioned trajectories ($m{=}4$ in the main benchmark; the
construction requires $P+\lfloor(E-P)/m\rfloor\ge k$).  At each MoE layer, a seeded
permutation selects a shared core $S_\ell$ of size $P$; the remaining experts
are split round-robin into disjoint private sets $V_{\ell,1},\dots,V_{\ell,m}$.
This construction constrains distributed computation rather than assuming
that individual experts correspond to distinct semantic skills.
Agent $a$ is eligible for
\begin{equation}
  \mathcal{E}_{\ell,a}=S_\ell\cup V_{\ell,a}, \qquad
  |\mathcal{E}_{\ell,a}|\approx P+\frac{E-P}{m},
  \label{eq:eligibility}
\end{equation}
and the routed logits become
\begin{equation}
  \widetilde r_{\ell,t,e}^{(a)}=
  \begin{cases}
    r_{\ell,t,e}, & e\in\mathcal{E}_{\ell,a},\\
    -\infty,      & e\notin\mathcal{E}_{\ell,a}.
  \end{cases}
  \label{eq:mask}
\end{equation}
The pretrained router still chooses its usual top-$k$ experts, now within the
eligible set of Equation~\eqref{eq:mask}.  Nothing is trained, and the mask
persists throughout the routed prefix.  At $P=E$, all trajectories reduce to
deterministic copies of the base model; decreasing $P$ increases private
capacity but risks excluding useful experts.  For \model, $E=256$, $k=8$, and
there are 40 MoE
layers~\citep{qwen36modelcard}. At the main benchmark setting $P=160$,
each trajectory can access 184 of 256 experts per layer; at $P=192$, it
can access 208.

The four trajectories share weights and execute along the batch dimension,
while retaining separate token histories and KV caches. A batched generation
request therefore contains multiple autoregressive forward steps over one
model instance.

The hybrid configuration limits constrained routing to a 2,000-token
prefix, after which unfinished trajectories continue with unrestricted
expert routing in the same frozen model. Appendix~\ref{app:handoff-frequency}
analyzes cap attainment and recorded handoffs in the saved runs.

\begin{figure}[!htbp]
  \centering
  \includegraphics[width=0.72\linewidth]{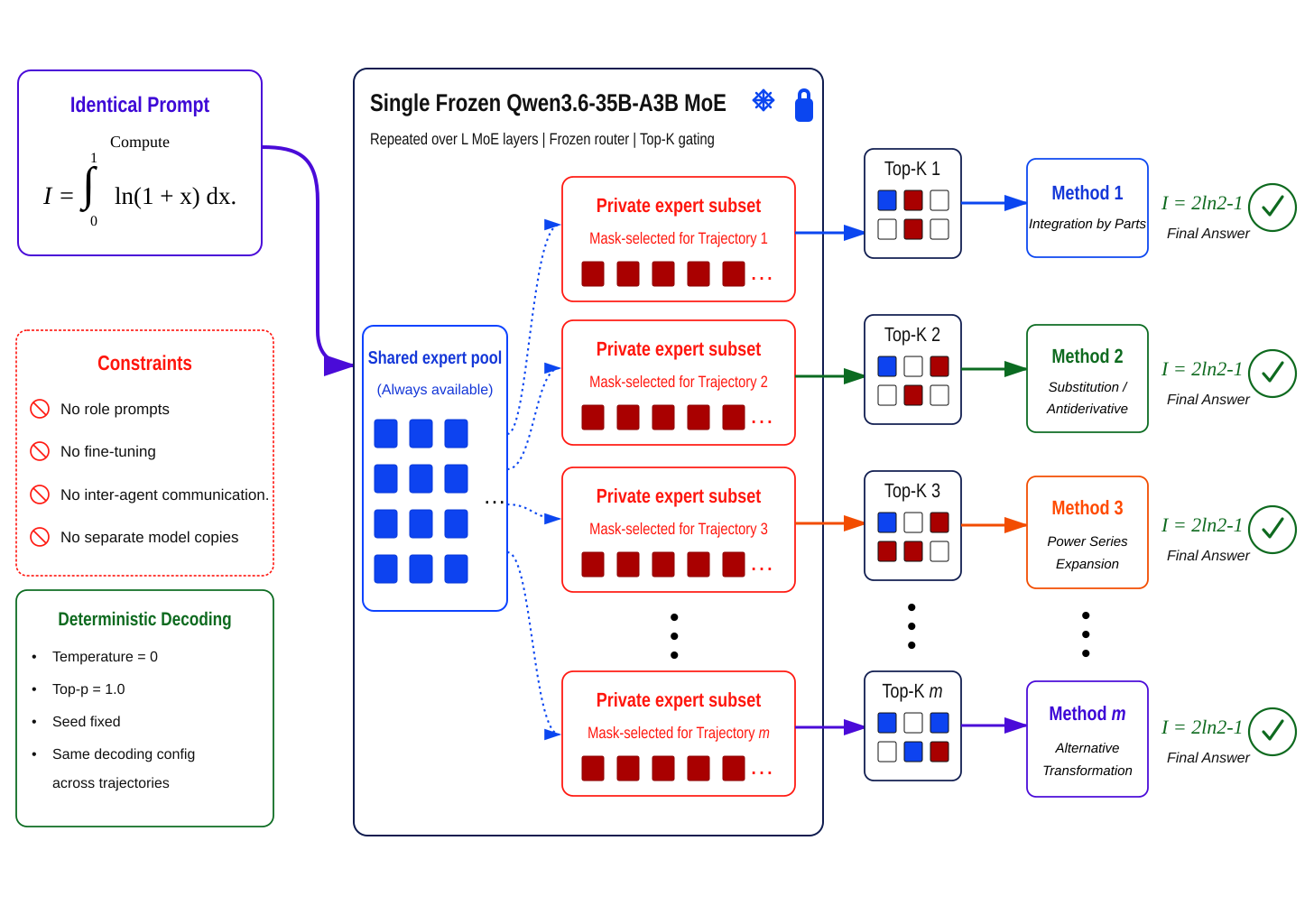}
  \caption{Architecture of \acr{}.  Four copies of an identical prompt
  share one frozen MoE backbone. Persistent shared/private masks give
  trajectories distinct expert eligibility while preserving the learned
  token router.}
  \label{fig:hybrid}
\end{figure}

\section{Experiments}
\label{sec:experiments}

\subsection{Tasks, model, and conditions}

\paragraph{Task and model.}
The main experiments use a frozen BF16 \model{} checkpoint, a 35B-parameter sparse
MoE with roughly 3B active parameters per
token~\citep{qwen36modelcard}. We use a synthetic 11-problem development set of
mathematics and physics derivations with verifiable answers and plausibly
multiple methods for configuration selection. We report these development
probes separately from the main evaluation, which uses a screened benchmark drawn
from a hand-authored 200-problem, 12-category candidate
pool spanning integrals, series, products, combinatorial identities,
probability, functional equations, number theory, inequalities, mechanics,
E\&M, thermodynamics, and quantum physics. Each problem was screened using six
temperature samples and retained only if the outputs showed multiple
solution paths and at least one correct solution; 126 problems passed.  The result
is a diagnostic set conditioned on evidence of multiple solution paths in
temperature-generated outputs. Our conclusions concern this selected
population; broader task coverage is discussed in Section~\ref{sec:limitations}.

\paragraph{Comparison methods and generation budget.}
All methods in the main benchmark use the same output budget: four generations per
problem and at most 4,096 new tokens each. This matches trajectory count
and output caps, not total compute or input-token usage.  \acr{} uses deterministic decoding and one
persistent mask per trajectory, sweeping $P\in\{96,160,192\}$ at benchmark
scale, with a repetition penalty of 1.05 and thinking disabled. The
baseline draws four i.i.d.\ temperature samples from the unpatched model
($T\in\{0.5,0.9,1.1\}$ at benchmark scale and on the development sweep).  The
Expert-Sample baseline applies routing noise~\citep{chen2026expertsample}
at benchmark scale: at every token and MoE layer, the top $k_{\mathrm{keep}}=5$
experts by gating logit are retained, the remaining three
of the $k=8$ selections are Gumbel-sampled from candidate ranks $6$--$32$
at $\tau{=}1.0$, and weights are renormalized from the
original global softmax, following the published defaults for this backbone class.
The RoE baseline applies
routing noise~\citep{zibakhsh2025roe} at benchmark scale: at every generated token the
selection becomes $\operatorname{TopK}(R+\tau G, k)$ with $G$ i.i.d.\
standard Gumbel on the middle MoE layers (the first and last two layers stay
clean), and $N{=}8$ perturbed proposals per token run in one batched forward pass
whose next-token distributions are probability-averaged into a single greedy
prediction, with only the clean path's KV cache retained.  The proposal
count and the fixed $\tau{=}0.5$ are compute-driven deviations from the
published per-model-tuned settings ($N{=}32/64$); decoding is greedy and
otherwise identical to \acr{}. We also compare against informed sequential
diversity prompting, where calls~2--4 repeat the problem with excerpts
of all prior solutions and an
instruction to use a fundamentally different method.  It uses the same
four-call budget, but later calls receive inter-attempt information and an
explicit method hint, and the sequential dependency precludes batching
(Table~\ref{tab:latency}, sequential arm).  All other methods share
one neutral problem prompt with no method hints.
The held-out saturation probe draws up to 50 temperature samples to test whether
a solution path merely has low sampling probability.

\subsection{Judging protocol}

\paragraph{Correctness.}
A trajectory is \emph{soft-correct} when it reaches the mathematically
correct answer in its reasoning, including when that answer is omitted
from the final box or affected by LaTeX formatting errors. Judges read the
complete raw prediction rather than relying on boxed-answer extraction.
We use ``correct methods'' for the mathematical routes in these
answer-correct trajectories; intermediate-step validity is a separate
criterion from final-answer correctness.

\paragraph{Method diversity.}
For problem $x$ and trajectory $i$, let $\mu_i(x)$ denote its executed
mathematical route and $C_i(x)$ indicate soft-correctness. Each trajectory
contributes at most one method. We define
\begin{equation}
  \mathrm{UCM}@m(x)=
  \left|\left\{\mu_i(x) \;:\; i\in\{1,\ldots,m\},\ C_i(x)=1\right\}\right|.
\end{equation}
Judges count substantive differences in mathematical steps as method
diversity, including differences within the same broad method family.
Rephrasing and reordering equivalent steps do not qualify. Thus the
comparison operates at the level of executed routes rather than only
named techniques. Multiple derivations within a single trajectory receive
at most one unit of credit; repeated routes across trajectories count once.
With four trajectories per problem, $0\leq\mathrm{UCM}@4(x)\leq4$.

\paragraph{Multi-method problem counts.}
We report the number of benchmark problems with at least two
distinct primary methods among soft-correct trajectories:
\begin{equation}
  N_{\geq2}=\sum_{x\in\mathcal{D}}
  \mathbf{1}\!\left[\mathrm{UCM}@4(x)\geq2\right].
\end{equation}
The normalized rate $N_{\geq2}/|\mathcal{D}|$ includes all 126 problems,
including those with no correct solution. Counts are computed within runs
and averaged across seeds, not unioned across runs. Mean UCM@4 and
$N_{\geq3}$ provide complementary measures of method count; per-trajectory soft-correctness is reported alongside diversity.
``Two solutions'' means two distinct primary methods, not repeated answers.

All model-judged arms use the same judging instructions (Appendix~\ref{app:judge-prompt}).
Judges are instructed to read raw JSON predictions in full and assess them
without consulting previous results, artifacts, judgments, or manuals.
The prompt requests pool-wise UCM@4 and counts of problems with at least
two and three distinct solutions. GLM 5.3 with max thinking and GPT-5.6 Sol
with high thinking provide separate model-based evaluations. Main-table
GLM results average three seeds; Sol results use the per-arm seed counts
in Table~\ref{tab:sol-review}. GLM and Sol are evaluators of generated solutions, not additional generation
backbones. Judge-specific analyses are reported separately; the human-audit
protocol and results appear in Section~\ref{sec:human-audit}.

\subsection{Resource accounting}
ACR stores one backbone and preserves the top-$k$ active-expert count.
Its FP32 masks require 160~KiB (40 layers, four agents, 256 experts);
a packed representation would require about 5~KiB. No adapters or extra
weights are trained. Expert-Sample likewise preserves the activation count;
RoE instead processes multiple proposals and temporarily retains their KV
rows before keeping only the clean path.

Table~\ref{tab:latency} reports a short-generation microbenchmark.
ACR remains close to batched temperature sampling in latency and peak
memory, while RoE and sequential prompting take longer. The 1,024-token
cap never reaches the 2,000-token handoff, so these measurements do not
characterize the full 4,096-token hybrid configuration. Backbone weights
dominate memory throughout.

\begin{table}[t]
  \centering
  \caption{Latency and VRAM at matched budget (four completions per problem,
  3 screened problems, max $1{,}024$ new tokens, single 80~GB A100, torch
  fallback, warmup excluded; means over problems).  ``Throughput'' counts
  total generated tokens divided by elapsed wall-clock time.}
  \label{tab:latency}
  \small
  \begin{tabular}{lcccc}
    \toprule
    Arm & Seconds/problem & Throughput (tok/s) & Peak alloc & Peak reserved \\
    \midrule
    Temperature, batched & 84.5 & 48.4 & 65.82~GB & 66.00~GB \\
    \acr{} ($P{=}160$) & 86.3 & 47.4 & 66.10~GB & 66.39~GB \\
    Expert-Sample & 95.2 & 43.0 & 65.82~GB & 66.00~GB \\
    RoE ($N{=}8$) & 417.0 & 9.8 & 66.46~GB & 66.99~GB \\
    Multi-agent, sequential & 323.6 & 12.7 & 65.52~GB & 65.58~GB \\
    \bottomrule
  \end{tabular}
\end{table}

\section{Results}
\label{sec:results}

We report the screened benchmark, blinded human audit, qualitative example,
and ablations. Development and held-out probes appear in
Appendix~\ref{sec:dev}.

\subsection{Benchmark scale: the screened 126}
\label{sec:benchmark}

The benchmark contains 126 screened problems. The method-diversity
evaluation covers \acr{} at its main setting $P{=}160$,
three temperature settings, both routing-noise controls, and the
informed sequential prompting control; the remaining pool sizes are ablated
in Section~\ref{sec:ablations}.

\begin{figure}[t]
  \centering
  \includegraphics[width=0.65\linewidth]{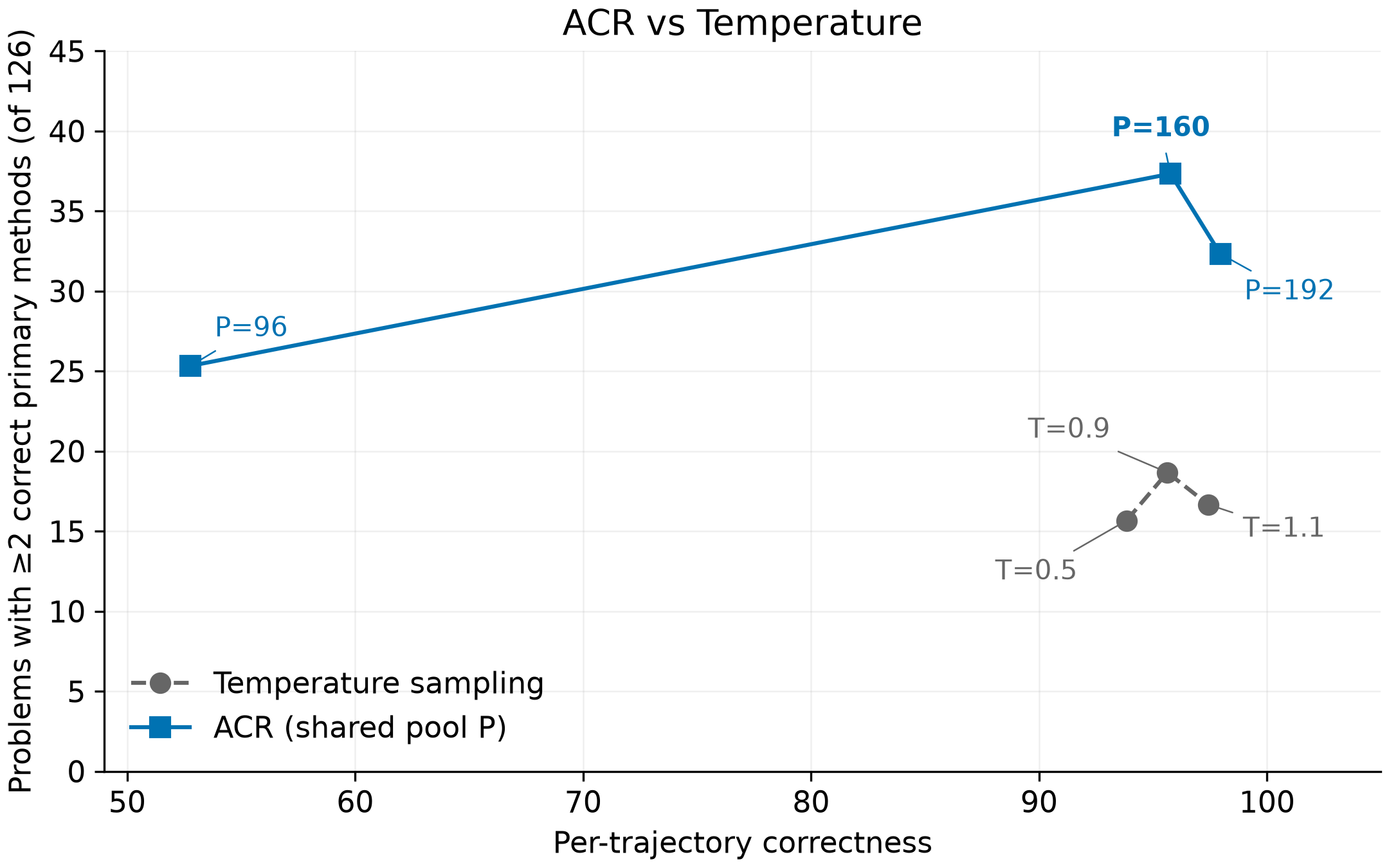}
  \caption{GLM multi-method problem counts versus correctness at matched output budgets. Points are three-seed means from Table~\ref{tab:bench-methods}; lines connect tested settings. ACR $P=160$ reaches 37.3 multi-method problems versus 18.7 for temperature $T=0.9$.}
  \label{fig:bench-pareto}
\end{figure}

\paragraph{Method diversity under GLM 5.3.}
At the main setting, ACR yields at least two correct primary methods on
37.3 of 126 problems, versus 18.7 for the strongest temperature setting on
this metric ($T=0.9$): approximately a twofold increase, averaged over three
seeds (Table~\ref{tab:bench-methods}). Correctness is 95.77\% versus 95.63\%.
The secondary mean UCM@4 is 1.323 versus 1.159. The approximately twofold
increase refers to the prevalence of problems with alternative methods.
The most diverse examples include order statistics, spin measurement, and
the alternating Basel sum. Figure~\ref{fig:bench-pareto} shows the
correctness--diversity comparison across pool sizes.

\paragraph{GPT-5.6 Sol evaluation.}
The GPT-5.6 Sol evaluation also favors ACR in multi-method problem counts:
45.00 problems with at least two correct methods versus 27.00 for the
strongest evaluated temperature setting ($T=0.9$), a 1.67-fold increase
(Table~\ref{tab:sol-review}). Both settings are averaged over three seeds. Correctness
is 97.69\% versus 100.00\%, and mean UCM@4 is 1.3915 versus 1.2328.
In both benchmark tables, bold threshold counts identify the best ACR
pool setting, excluding the separately listed informed comparator.

\begin{table}[t]
  \centering
  \caption{GPT-5.6 Sol benchmark: 126 problems, four trajectories per
  problem. Values are averaged over the listed seeds. $N_{\geq2}$ and
  $N_{\geq3}$ count problems.}
  \label{tab:sol-review}
  \footnotesize
  \setlength{\tabcolsep}{3pt}
  \begin{tabular}{lrrrrr}
    \toprule
    Arm & Seeds & $N_{\geq2}$ & $N_{\geq3}$ & Correct (\%) & UCM@4 \\
    \midrule
    Temp. $T=0.5$ & 2 & 25.00 & 1.50 & 100.00 & 1.2103 \\
    Temp. $T=0.9$ & 3 & 27.00 & 2.33 & 100.00 & 1.2328 \\
    Expert-Sample & 1 & 21.00 & 2.00 & 100.00 & 1.1825 \\
    RoE & 1 & 12.00 & 1.00 & 100.00 & 1.1032 \\
    \midrule
    \acr{} ($P=96$) & 1 & 24.00 & 2.00 & 60.71 & 1.0873 \\
    \acr{} ($P=160$) & 3 & \textbf{45.00} & \textbf{4.00} & 97.69 & 1.3915 \\
    \acr{} ($P=192$) & 3 & 37.67 & 2.67 & 98.94 & 1.3201 \\
    \midrule
    DivPrompt (informed, seq.) & 1 & 125.00 & 124.00 & 96.03 & 3.7222 \\
    \bottomrule
  \end{tabular}
\end{table}

\begin{table}[t]
  \centering
  \caption{GLM 5.3 benchmark: counts of
  problems with at least two correct methods, correctness, and secondary
  diversity metrics on the screened 126-problem benchmark. All results are averaged over
  three seeds.}
  \label{tab:bench-methods}
  \small
  \begin{tabular}{lcccc}
    \toprule
    Arm & $N_{\geq2}$ & $N_{\geq3}$ & Correct (\%) & Mean UCM@4 \\
    \midrule
    Temp. $T{=}0.5$ & 15.67 & 0.67 & 93.84 & 1.130 \\
    Temp. $T{=}0.9$ & 18.67 & 1.33 & 95.63 & 1.159 \\
    Temp. $T{=}1.1$ & 16.67 & 0.33 & 97.42 & 1.150 \\
    Expert-Sample & 14.33 & 1.00 & 99.80 & 1.127 \\
    RoE & 17.67 & 0.00 & 99.40 & 1.140 \\
    \midrule
    \acr{} ($P{=}96$) & 25.33 & 3.00 & 52.78 & 1.248 \\
    \acr{} ($P{=}160$) & \textbf{37.33} & \textbf{4.33} & 95.77 & 1.323 \\
    \acr{} ($P{=}192$) & 32.33 & 3.67 & 97.95 & 1.280 \\
    \midrule
    DivPrompt (informed, seq.) & 124.33 & 119.00 & 94.44 & 3.571 \\
    \bottomrule
  \end{tabular}
\end{table}

\paragraph{Paired diversity comparisons.}
For GPT-5.6 Sol, ACR ($P=160$, three seeds) yields at least two methods on
35.71\% of problems versus 21.43\% for temperature $T=0.9$, a gain of
14.29 percentage points (paired-bootstrap 95\% CI: $[8.20,20.37]$;
sign-flip $p=1.49\times10^{-5}$). Intervals favor ACR over each listed
blind comparator and favor informed DivPrompt over ACR
(Table~\ref{tab:paired-rates}). Paired mean-UCM@4 comparisons show the same
direction (Appendix~\ref{app:ucm-statistics}). Mask-structure comparisons
remain descriptive.

\begin{table}[htbp]
  \centering
  \caption{Paired comparisons of the rate of problems with at least two
  methods under GPT-5.6 Sol. ACR ($P=160$) has a 35.71\% rate;
  seed counts are given in Table~\ref{tab:sol-review}. Differences and
  paired-bootstrap 95\% CIs are in percentage points; $\Delta$ is ACR minus
  comparator. The final column reports sign-flip $p$-values.}
  \label{tab:paired-rates}
  \small
  \begin{tabular}{lrrrr}
    \toprule
    Comparator & Rate (\%) & $\Delta$ (pp) & 95\% CI (pp) & $p$ \\
    \midrule
    RoE & 9.52 & +26.19 & $[19.58,\,32.80]$ & $3.74\times10^{-12}$ \\
    Expert-Sample & 16.67 & +19.05 & $[11.64,\,26.46]$ & $1.93\times10^{-6}$ \\
    DivPrompt & 99.21 & $-63.49$ & $[-69.84,\,-56.88]$ & $1.23\times10^{-32}$ \\
    Temperature $T=0.5$ & 19.84 & +15.87 & $[9.92,\,21.96]$ & $6.75\times10^{-7}$ \\
    Temperature $T=0.9$ & 21.43 & +14.29 & $[8.20,\,20.37]$ & $1.49\times10^{-5}$ \\
    \bottomrule
  \end{tabular}
\end{table}

\paragraph{Routing-noise controls.}
Expert-Sample and RoE produce fewer multi-method problems despite high
correctness (Table~\ref{tab:bench-methods}). On spin measurement, all eight
baseline trajectories use operator-eigenvector derivations; rotation-operator
checks add no primary-method credit. This comparison changes both eligibility
and structure, so it does not isolate persistence.

\paragraph{Informed sequential prompting.}
Showing each later call prior solutions and requesting a different method
yields the highest GLM diversity (Table~\ref{tab:bench-methods}) and finds routes absent from every blind arm.
This protocol can also be automated, but its later calls depend on earlier
solutions and explicit requests for alternatives. ACR generates its trajectories
independently from one neutral prompt. Among the tested blind, batched,
prompt-matched arms, it yields the most problems with at least two correct
methods, requiring no user-assigned roles or prescribed techniques.

\subsection{Blinded human audit}
\label{sec:human-audit}

Four human experts evaluated outputs across three seeds on a 30-problem
subset selected by stratified random sampling to preserve the original
dataset's composition. Model and setting identities were hidden.
Table~\ref{tab:human-audit} reports method counts and correctness separately
from the model-judged full-benchmark results.

Qwen ACR $P=160$ yields at least two methods on 12.0 problems versus 9.0
for temperature sampling, a 1.33-fold increase, with correctness of 96.4\%
versus 98.3\%. Gemma ACR $P=80$ yields 14.0 versus 7.0, a twofold increase,
with correctness of 86.2\% versus 100.0\%. These descriptive subset results
support the direction of the diversification gains on both backbones while
showing a correctness tradeoff. Qwen $P=96$ has the largest three-method
count but substantially lower correctness, reinforcing the need to report
both outcomes.

\begin{table}[htbp]
  \centering
  \caption{Blinded human audit on 30 problems. Threshold columns report
  mean problem counts out of 30 across three seeds; UCM@4 reports mean
  method counts. Four experts evaluated outputs with model and setting
  identities hidden.}
  \label{tab:human-audit}
  \small
  \begin{tabular}{lrrrr}
    \toprule
    Setting & $N_{\geq2}$ & $N_{\geq3}$ & Correctness (\%) & Mean UCM@4 \\
    \midrule
    Qwen temperature & 9.0 & 1.0 & 98.3 & 1.33 \\
    Expert-Sample & 6.0 & 2.0 & 99.2 & 1.27 \\
    RoE & 4.0 & 0.0 & 99.2 & 1.13 \\
    Qwen \acr{} ($P=96$) & 10.0 & \textbf{5.0} & 57.5 & 1.33 \\
    Qwen \acr{} ($P=160$) & \textbf{12.0} & 2.3 & 96.4 & 1.51 \\
    Qwen \acr{} ($P=192$) & 9.7 & 2.7 & 97.5 & 1.43 \\
    \midrule
    Gemma baseline & 7.0 & 1.0 & 100.0 & 1.27 \\
    Gemma \acr{} ($P=80$) & \textbf{14.0} & \textbf{4.67} & 86.2 & 1.65 \\
    \bottomrule
  \end{tabular}
\end{table}

\subsection{Reading the derivations: rolling down an incline}
\label{sec:cases}

\noindent\textbf{Problem.} \emph{A rigid body with moment of inertia
$I=\kappa mR^2$ rolls without slipping down an incline at angle $\alpha$.
Find its center-of-mass acceleration and provide a complete derivation.}

All eight trajectories in the selected runs reach
$a=g\sin\alpha/(1+\kappa)$. Their derivations differ:
ACR includes a constrained Lagrangian derivation and explicit energy checks,
while all four temperature samples use force and torque balance
(Table~\ref{tab:rolling-routes}). Three approaches occur across ACR's full
texts, with energy appearing as a secondary derivation; the primary routes
are Lagrangian and Newtonian. Appendix~\ref{app:rolling-case} records the
trajectory mapping. Appendix~\ref{app:pie-case} renders a seed-0 inclusion--exclusion comparison; development examples appear in Appendix~\ref{app:dev-cases}.

\begin{table}[t]
  \centering
  \caption{Rolling down an incline: ACR $P=160$ versus
  temperature $T=0.9$, four trajectories each. Equations summarize the raw
  derivations with normalized notation; secondary checks receive no extra
  primary-method credit.}
  \label{tab:rolling-routes}
  \footnotesize
  \setlength{\tabcolsep}{4pt}
  \begin{tabular}{lp{0.40\linewidth}ll}
    \toprule
    Route & Distinguishing equations & ACR & Temperature \\
    \midrule
    Newtonian & $mg\sin\alpha-f=ma$, $fR=I\ddot\theta$, $a=R\ddot\theta$
      & Primary: 2--4 & Primary: 1--4 \\
    Lagrangian & $mg\sin\alpha-ma+\lambda=0$, $I\ddot\theta+R\lambda=0$
      & Primary: 1 & Absent \\
    Energy & $\frac{d}{dt}[\frac12m(1+\kappa)v^2-mgx\sin\alpha]=0$
      & Check: 1, 2 & Absent \\
    \bottomrule
  \end{tabular}
\end{table}

\subsection{Ablation studies}
\label{sec:ablations}

\paragraph{Persistence and mask structure.}
We fix eligibility at 184 experts per trajectory and layer, matching
$P=160$, and retain the same handoff and generation settings.
Token-resampled ACR redraws the shared/private assignment at each decoding
step during the routed prefix, isolating persistence.
Independent-persistent masks sample 184 experts per trajectory and layer
without a common core or disjoint private sets. This tests the shared/private
construction as a whole, including its effect on overlap between trajectories.

\begin{table}[htbp]
  \centering
  \caption{Persistence and mask structure at matched eligibility: 184 experts per trajectory per layer, matching ACR $P=160$. Threshold columns are mean problem counts out of 126.}
  \label{tab:persistence}
  \small
  \begin{tabular}{lcccc}
    \toprule
    Arm & $N_{\geq2}$ & $N_{\geq3}$ & Correct (\%) & Mean UCM@4 \\
    \midrule
    Independent-persistent & 39.67 & 3.67 & 96.49 & 1.3212 \\
    Token-resampled ACR & 42.33 & 3.33 & 99.21 & 1.3574 \\
    Hybrid \acr{} & \textbf{45.00} & \textbf{4.00} & 97.69 & 1.3915 \\
    \bottomrule
  \end{tabular}
\end{table}

ACR yields multi-method solutions on 5.33 more problems than independent
masks and 2.67 more than resampled masks (Table~\ref{tab:persistence}).
It improves correctness over independent masks but sacrifices 1.52 percentage
points relative to resampling. Both controls retain much of the diversification
effect; the full construction leads on diversity by modest descriptive margins.

\paragraph{Matched handoff comparison.}
On 20 problems at $P=160$, switching yields eight additional
end-of-sequence (EOS) completions and one additional Sol-judged correct
trajectory, with unchanged aggregate
diversity (Table~\ref{tab:paired-handoff}). Only 28 of 80 trajectories
remain unfinished at the boundary. Appendix~\ref{app:paired-handoff} details
the matched-prefix protocol; Appendix~\ref{app:handoff-frequency} reports
full-benchmark handoff frequencies.

\begin{table}[htbp]
\centering
\caption{Matched handoff comparison on 20 problems at $P=160$, mask seed 0.
Correctness and method diversity are judged by GPT-5.6 Sol; EOS completion
is recorded directly from generated token IDs.}
\label{tab:paired-handoff}
\small
\setlength{\tabcolsep}{3pt}
\begin{tabular}{lccccc}
\toprule
Arm & \shortstack{Problems\\$\geq2$ methods} & \shortstack{Problems\\$\geq3$ methods} & \shortstack{Mean\\UCM@4} & Correctness & Reached EOS \\
\midrule
$\mathrm{Switch}_{2000}$ & 17/20 & 3/20 & 2.00 & 93.75\% & 92.5\% \\
No switch & 17/20 & 3/20 & 2.00 & 92.50\% & 82.5\% \\
\bottomrule
\end{tabular}
\end{table}

\paragraph{Pool size $P$.}
Both model judges rank $P=160$ above $P=192$ and $P=96$ in diversity
(Tables~\ref{tab:bench-methods} and~\ref{tab:sol-review}). Small pools can
induce looping; larger pools restore correctness but reduce diversity.

\begin{table}[htbp]
  \centering
  \caption{Backbone ablation: Gemma-4-26B-A4B (128 experts, top-8), one
  seed, 126 problems, four trajectories; $P{=}80$ matches Qwen $P{=}160$'s
  shared-core fraction.}
  \label{tab:gemma-backbone}
  \footnotesize
  \begin{tabular}{lcccc}
    \toprule
    Arm & $N_{\geq2}$ & $N_{\geq3}$ & Correct (\%) & Mean UCM@4 \\
    \midrule
    Temperature $T{=}1.1$ & 21 & 3 & 87.6 & 1.19 \\
    \acr{} ($P{=}80$) & \textbf{52} & \textbf{11} & 84.9 & 1.49 \\
    \bottomrule
  \end{tabular}
\end{table}

\paragraph{Backbone.}
On Gemma-4-26B-A4B, the fraction-matched $P=80$ setting yields 52 versus
21 multi-method problems, with a 2.7-percentage-point correctness reduction
(Table~\ref{tab:gemma-backbone}). This one-seed comparison uses the same
budget and protocol. With 128 experts, Gemma gives each trajectory 12
private experts, versus 24 for Qwen.

\subsection{Limitations and future work}
\label{sec:limitations}

Masks constrain computation without assigning semantic specializations
and may yield the same method~\citep{wang2026illusion}. Matched-eligibility
contrasts remain descriptive; paired uncertainty and expert assessment of
intermediate-step validity are outside this evaluation. UCM@4 and
$N_{\geq3}$ complement the two-method threshold.

The screened benchmark targets problems with multiple solution methods.
The human audit covers 30 problems; agreement between reviewers is not
quantified. Broader task and model coverage,
layer-specific masks, resource scaling, and downstream selection or
aggregation remain future work.

\section{Conclusion}
\label{sec:conclusion}

Expert eligibility provides an inference-time control for reasoning diversity
within one frozen MoE, without user-assigned roles or prescribed methods.
At matched output budgets, ACR yields at least two correct primary methods
on approximately twice as many problems as the strongest temperature setting
under GLM and 1.67 times as many under GPT-5.6 Sol. A blinded four-expert
audit supports the direction of these gains on both backbones, with lower
correctness for ACR. The matched-eligibility
controls retain much of the diversification effect, with modest advantages
for the full construction. Informed sequential prompting achieves greater
diversity using previous attempts; ACR explores alternatives independently
in one batched request.

\clearpage
\section*{AI Statement}
AI tools were used to polish the manuscript's language and presentation,
generate the synthetic 11-problem development set, and serve as judges for
solution evaluation. The model-based evaluations are distinguished from
the blinded human audit in the evaluation protocol and results.

\bibliographystyle{iclr2027_conference}
\bibliography{resource_aware_acr_2026}

\clearpage
\appendix
\section{LLM judging prompt}
\label{app:judge-prompt}
The following prompt was used for the LLM evaluations, reproduced with
its original wording. ``Manually'' instructs the model to read the raw
predictions directly; the evaluations are model judgments.

\begin{quote}
read raw predictions of jsons manually fully. judge how many unique correct methods are predicted for each problem. its fine if the solutions are soft-correct basically ignoring latex misformating and if the answer is reached in the reasoning but missed in boxed. dont look at previous results/artifacts/judges/manuals. this should be purely your effort. also if there is diversity in the steps in a particular method consider it as method diversity. in the end print pool-wise UCM@4 and number of problems with atleast 2 solutions and number of problems with alteast 3 solutions. Each solution can have only 1 trajectory. Which implies each problem can have atmost 4 distinct trajectories. Rephrasing or reordering of steps is not method diversity.
\end{quote}

\section{Paired mean-UCM@4 comparisons}
\label{app:ucm-statistics}
The secondary mean-UCM@4 comparisons use ACR at $P=160$ over three seeds.
Paired-bootstrap intervals exclude zero in favor of ACR against the listed
blind comparators, and in favor of informed DivPrompt against ACR.

\begin{table}[htbp]
  \centering
  \caption{Paired UCM@4 comparisons. $\Delta$ is ACR minus comparator;
  ACR uses $P=160$ over three seeds, with mean UCM@4 approximately
  1.3915. Point differences use the displayed means. Intervals are paired-bootstrap
  95\% CIs; the final column reports sign-flip $p$-values.}
  \label{tab:paired-ucm}
  \small
  \begin{tabular}{lrrrr}
    \toprule
    Comparator & UCM@4 & $\Delta$ & 95\% CI & $p$ \\
    \midrule
    RoE & 1.1032 & +0.2883 & $[0.2196,\,0.3598]$ & $3.56\times10^{-13}$ \\
    Expert-Sample & 1.1825 & +0.2090 & $[0.1243,\,0.2910]$ & $3.59\times10^{-6}$ \\
    DivPrompt & 3.7222 & $-2.3307$ & $[-2.4312,\,-2.2275]$ & $4.70\times10^{-38}$ \\
    Temperature $T=0.5$ & 1.2103 & +0.1812 & $[0.1164,\,0.2474]$ & $2.30\times10^{-7}$ \\
    Temperature $T=0.9$ & 1.2328 & +0.1587 & $[0.0899,\,0.2302]$ & $2.59\times10^{-5}$ \\
    \bottomrule
  \end{tabular}
\end{table}

\section{Rolling-incline trajectory comparison}
\label{app:rolling-case}
The selected ACR run uses $P=160$ and mask seed 2; the temperature run uses
$T=0.9$, top-$p=0.95$, and seed 0. Both generate four trajectories per
problem with a 4,096-token cap. Trajectory numbers follow the order of
raw predictions in each source file.

ACR trajectory 1 constructs $L=\frac12m\dot x^2+
\frac12I\dot\theta^2+mgx\sin\alpha$, imposes $x-R\theta=0$ with a
multiplier, and eliminates that multiplier to obtain the acceleration.
It then checks the result by differentiating the conserved energy.
Trajectory 2 first eliminates friction from the translational and rotational
equations, then independently derives the answer from energy conservation.
Its second heading says ``Lagrangian Mechanics (Energy Conservation)'',
but the executed calculation differentiates energy rather than applying
Euler--Lagrange equations. Trajectory 3 uses force and torque balance,
followed by an effective-mass check; trajectory 4 uses force and torque
balance. Each temperature trajectory eliminates friction using the same
Newtonian equations. Thus route labels follow the executed equations,
not the model's headings. This selected example illustrates route coverage;
it is not an additional aggregate benchmark result.

\section{Development and held-out probes}
\label{sec:dev}

\begin{figure}[t]
  \centering
  \begin{subfigure}[t]{0.48\linewidth}
    \centering
    \includegraphics[width=\linewidth]{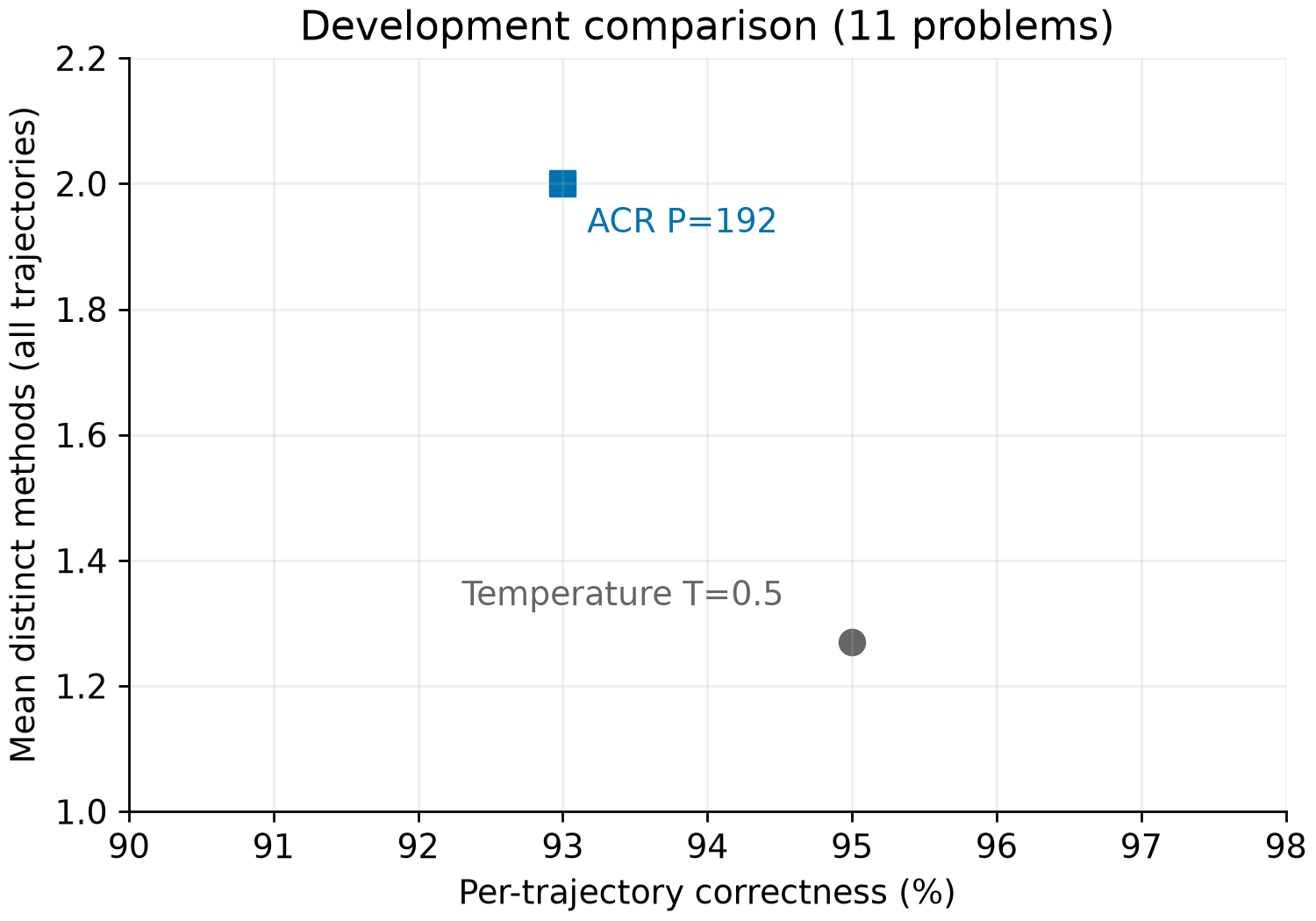}
    \caption{Development comparison.}
    \label{fig:dev-pareto}
  \end{subfigure}\hfill
  \begin{subfigure}[t]{0.48\linewidth}
    \centering
    \includegraphics[width=\linewidth]{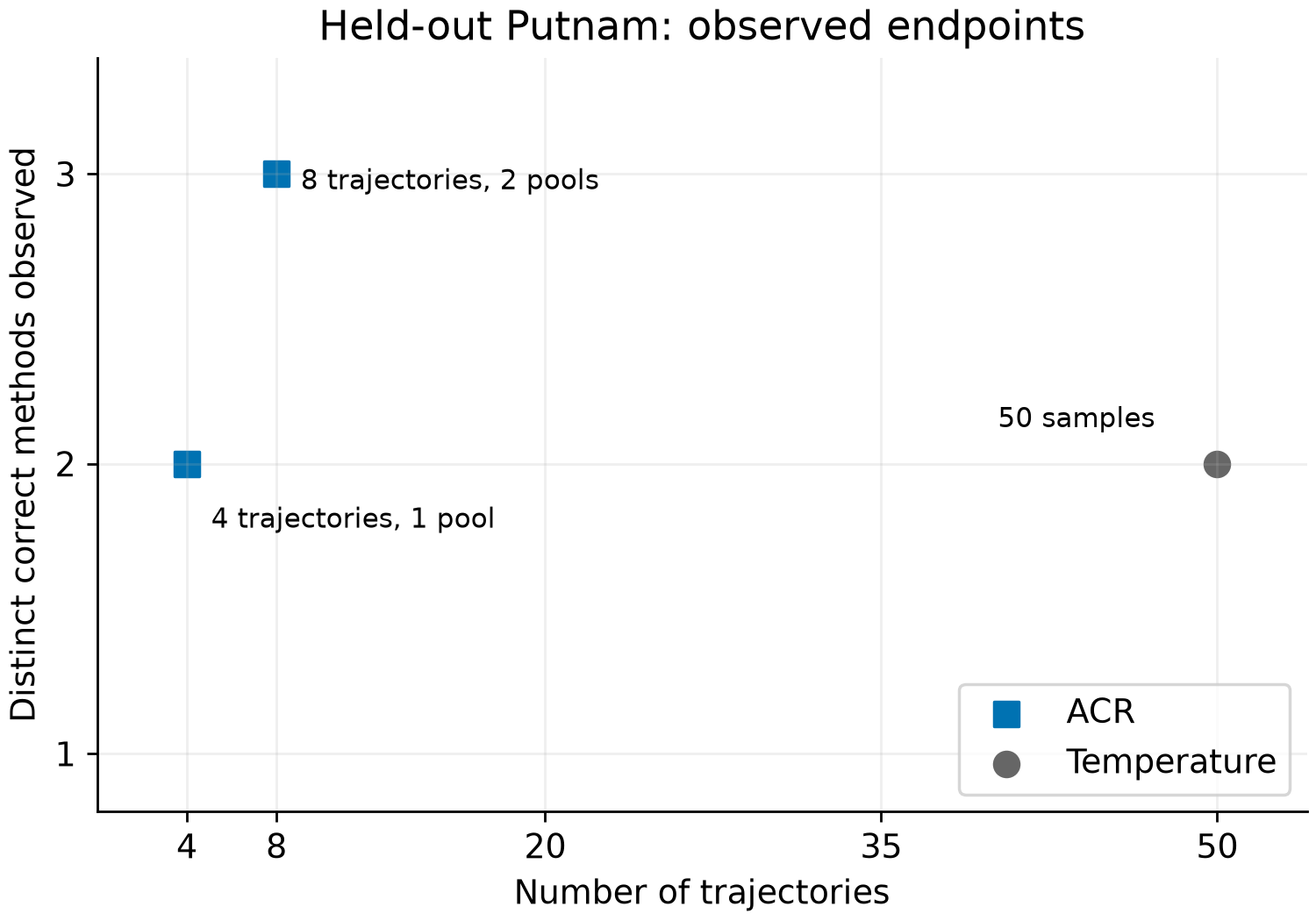}
    \caption{Held-out Putnam saturation.}
    \label{fig:putnam}
  \end{subfigure}
  \caption{Development and held-out probes. \textbf{(a)} Reported development
  comparison; method counts include all trajectories.  \textbf{(b)} On a held-out
  Putnam problem, 50 temperature samples yield two observed methods while eight
  routed trajectories expose a third.}
  \label{fig:diversity-probes}
\end{figure}

On the 11-problem development set, the reported $P=192$ configuration
reaches 93\% soft-correctness and 2.00 distinct methods per problem, compared
with 95\% and 1.27 for temperature $T=0.5$
(Figure~\ref{fig:dev-pareto}). These development method counts include all
trajectories and are separate from the correctness-filtered benchmark UCM@4.
The development set was used for configuration selection.

\section{Development case studies}
\label{app:dev-cases}
On $\int_0^\infty\ln(x)/(1+x^2)\,dx$, routed trajectories use reciprocal
symmetry, a series derivation, Mellin parameter differentiation, and the
$x=e^t$ odd-function transformation; four temperature samples repeat the
reciprocal route. For Tsirelson's bound, ACR produces operator-norm,
Choi--Jamio{\l}kowski, tensor-Pauli, and Bloch-sphere proofs, versus the
temperature baseline's operator-norm route. For Basel, $x^2$ cosine-series
and $x$ sine-series/Parseval derivations count separately only when used as
primary derivations. These examples motivate full-text method labeling.

\section{Inclusion--exclusion: rendered seed-0 solutions}
\label{app:pie-case}

\noindent\textbf{Problem.} \emph{State and prove inclusion--exclusion for
$|A\cup B\cup C|$, where $A$, $B$, and $C$ are finite sets.}

All four ACR trajectories ($P=160$, mask seed 0) and all four temperature
trajectories ($T=0.9$, sampling seed 0) reach the identity
\begin{equation}
\begin{aligned}
|A\cup B\cup C|={}&|A|+|B|+|C|\\
 &-|A\cap B|-|A\cap C|-|B\cap C|+|A\cap B\cap C|.
\end{aligned}
\label{eq:pie-case}
\end{equation}
The following are condensed renderings of completed arguments in the raw
outputs, with notation standardized; they are not verbatim transcripts.
Abandoned intermediate steps are omitted. Trajectory indices follow the
stored output order.

\paragraph{Element-by-element counting: ACR 1--2; temperature 1--4.}
For each $x\in A\cup B\cup C$, count its contribution to the right-hand
side of Equation~\eqref{eq:pie-case}. If it belongs to exactly one set,
its contribution is $1$. If it belongs to exactly two sets, its contribution
is $2-1=1$. If it belongs to all three, its contribution is $3-3+1=1$.
Thus each element of the union contributes exactly once. Summing these
contributions proves the formula. Writing the same membership cases with
indicator functions does not create a different route.

\paragraph{Disjoint-region decomposition: ACR 3.}
Let $r_A,r_B,r_C$ count elements in only $A$, only $B$, and only $C$;
let $r_{AB},r_{AC},r_{BC}$ count elements in exactly the indicated pair;
and let $r_{ABC}=|A\cap B\cap C|$. These seven regions partition the union.
Expanding the individual and intersection cardinalities gives
\begin{align*}
S_1&=|A|+|B|+|C|\\
   &=r_A+r_B+r_C+2(r_{AB}+r_{AC}+r_{BC})+3r_{ABC},\\
S_2&=|A\cap B|+|A\cap C|+|B\cap C|\\
   &=r_{AB}+r_{AC}+r_{BC}+3r_{ABC}.
\end{align*}
Consequently $S_1-S_2+r_{ABC}$ is the sum of the seven disjoint region
sizes, which equals $|A\cup B\cup C|$. This is the final region calculation
in trajectory 3 after it revisits the triple-intersection coefficient.

\paragraph{Reduction to the two-set identity: ACR 4.}
First establish $|X\cup Y|=|X|+|Y|-|X\cap Y|$ by checking the four
membership possibilities for an element. Apply this identity to $A\cup B$
and $C$:
\[
|A\cup B\cup C|=|A\cup B|+|C|-|(A\cup B)\cap C|.
\]
Distributivity gives $(A\cup B)\cap C=(A\cap C)\cup(B\cap C)$.
A second application of the two-set identity yields
\[
|(A\cup B)\cap C|=|A\cap C|+|B\cap C|-|A\cap B\cap C|.
\]
Substituting this and $|A\cup B|=|A|+|B|-|A\cap B|$ into the first
line proves Equation~\eqref{eq:pie-case}.

\paragraph{Primary routes and additional checks.}
The three route descriptions distinguish the executed steps under the
paper's rubric: pointwise multiplicity counting, explicit region-size
algebra, and recursive use of a two-set identity. ACR trajectories 1--2
complete the first route, trajectory 3 completes the second, and trajectory 4
completes the third. The four baseline trajectories complete the first
route. ACR trajectory 2 and baseline trajectory 4 also provide a two-set
proof after their first derivations; ACR trajectory 3 additionally expands
$1-(1-\mathbf{1}_A)(1-\mathbf{1}_B)(1-\mathbf{1}_C)$ as an indicator check.
These later arguments receive no extra primary-method credit. This
selected comparison illustrates proof structure, not a new aggregate result.

\section{Frequency of the 2,000-token handoff}
\label{app:handoff-frequency}
We analyze saved generation records without additional model inference.
Each run contains 126 problems and four trajectories per problem.
The $P=192$, seed-2 run records the number of prefix tokens and an explicit
switch flag for each trajectory. These flags identify 101 switched
trajectories (20.04\%) on 48 problems (38.10\%).

Earlier runs retain the decoded prefix length in characters but do not
store token counts or switch flags. We recover each prefix by slicing the
raw text at that character offset and retokenize it with the generation
model's tokenizer, without adding special tokens. Since decoding removes
special tokens and retokenization can change token boundaries, these are
estimates of cap attainment, not exact switch counts. Table~\ref{tab:handoff-frequency}
uses at least 1,995 retokenized tokens as the near-cap criterion.

\begin{table}[htbp]
\centering
\caption{Saved-output handoff analysis. Counts are trajectories out of 504
and problems out of 126 with at least one affected trajectory.
Estimated rows measure near-cap prefixes; the recorded row uses explicit
switch flags.}
\label{tab:handoff-frequency}
\small
\begin{tabular}{lrlrr}
\toprule
Pool $P$ & Seed & Evidence & Trajectories & Problems \\
\midrule
96  & 0 & Estimated & 276 & 103 \\
160 & 0 & Estimated & 167 & 73 \\
160 & 1 & Estimated & 66 & 43 \\
160 & 2 & Estimated & 149 & 67 \\
192 & 0 & Estimated & 123 & 56 \\
192 & 1 & Estimated & 48 & 28 \\
192 & 2 & Recorded & 101 & 48 \\
\bottomrule
\end{tabular}
\end{table}

At $P=160$, the near-cap criterion identifies 382 of 1,512 trajectories
(25.26\%) and 183 of 378 problem-runs (48.41\%) across three seeds.
Problem-runs count each problem separately in each seed, rather than taking
a union of problem IDs. A threshold of 1,990 tokens gives the same counts;
requiring at least 2,000 retokenized tokens gives 373 trajectories (24.67\%)
and 180 problem-runs (47.62\%). For the run with explicit flags, the
1,995-token estimate agrees with the recorded counts.

The historical records also contain appended text after prefixes well
below the cap, so the presence of continuation text alone is not used to
identify cap-triggered handoffs. This analysis measures observed switch
activation or estimated prefix-cap attainment. It does not measure looping,
correctness gains from switching, or whether a trajectory would have
succeeded under continued constrained routing.

\section{Matched switch-versus-no-switch evaluation}
\label{app:paired-handoff}
We select 20 problems from the screened benchmark using subset seed
20260926: one random problem from each of the 12 categories, followed by
eight problems sampled uniformly from the remainder. Selection uses only problem
IDs, without consulting outcomes. Both conditions use $P=160$, mask seed 0,
four trajectories, greedy decoding, repetition penalty 1.05, thinking
disabled, and a 4,096-token output cap.

Each pair shares the same masked prefix of up to 2,000 tokens.
Trajectories that reach EOS in this prefix are retained unchanged in both
conditions. Unfinished trajectories receive up to 2,096 further tokens,
with either unrestricted routing or the original eligibility mask. Both
conditions rebuild the prefix cache under their respective routing settings
and use the same active trajectory batch; execution order alternates by
problem. Thus the comparison controls restart mechanics while varying
routing eligibility during continuation.

Of the 80 trajectories, 52 reach EOS before or at the prefix boundary,
while 28 remain unfinished on 10 of the 20 problems. Among these 28 paired
continuations, 22 reach EOS with the switch versus 14 without it.
The repetition heuristic flags three full trajectories in the switch
condition and four in the no-switch condition. These diagnostics distinguish
termination and repetition from mathematical correctness: a trajectory may
reach a correct answer before exhausting its token budget without emitting
EOS. The observed gain is therefore eight additional EOS completions and
one additional soft-correct trajectory, with equal aggregate diversity
metrics. This subset supports a completion benefit; it does not establish
that unrestricted routing is universally necessary for correctness.

\end{document}